\documentclass{article} 
\usepackage{nips10submit_e,times}
\usepackage{amsmath, amssymb}
\usepackage{graphicx}

\def\eop {{\noindent\framebox[0.5em]{\rule[0.25ex]{0em}{0.75ex}}}}
\def\be {\begin{equation}}
\def\ee {\end{equation}}
\def\bea {\begin{eqnarray*}}
\def\eea {\end{eqnarray*}}
\def\beas {\begin{eqnarray*}}
\def\eeas {\end{eqnarray*}}
\newtheorem{theorem}{Theorem}
\newtheorem{lemma}{Lemma}
\newtheorem{claim}{Claim}
\newtheorem{claim-ap}{Claim}
\newcommand{\bfb}{{\bf b}}
\newcommand{\bfd}{{\bf d}}
\newcommand{\bfe}{{\bf e}}
\newcommand{\bfp}{{\bf p}}

\newcommand{\bfz}{{\bf z}}

\renewcommand{\S}{{\cal S}}

\newcommand{\Y}{{\cal Y}}

\newcommand{\bftheta}{\boldsymbol{\theta}}
\newcommand{\bflambda}{\boldsymbol{\lambda}}
\newcommand{\bfmu}{\boldsymbol{\mu}}

\title{Approximated Structured Prediction for Learning Large Scale Graphical Models}

\author{
Tamir Hazan \\
TTI Chicago\\
\texttt{hazan@ttic.edu} \\
\And
Raquel Urtasun \\
TTI Chicago \\
\texttt{rurtasun@ttic.edu} \\
}

\nipsfinalcopy

\begin{document}

\maketitle

This manuscript contains the proofs for "A Primal-Dual Message-Passing Algorithm for Approximated Large Scale Structured Prediction"

\begin{claim}
\label{claim:dual}
The dual program of the structured prediction program in (3) takes the form
$$
\max_{p_{x,y}(\hat y) \in \Delta_{\Y}} \sum_{(x,y) \in \S} \left(\epsilon H(\bfp_{x,y}) + \bfp_{x,y}^\top \bfe_y \right) - \frac{C^{1-q}}{q} \left\| \sum_{(x,y) \in \S} \sum_{\hat y \in Y} p_{x,y}(\hat y)\Phi(x,\hat y) - \bfd \right\|_q^q, 
$$
where $\Delta_{\Y}$ is the probability simplex over $\Y$ and $H(\bfp_{x,y}) = -\sum_{\hat y} p_{x,y}(\hat y) \ln p_{x,y}(\hat y)$ is the entropy.
\end{claim}
{\bf Proof:} We first describe an equivalent program to the one in (3) by adding variables $\mu(x,\hat y)$ instead of $ \bftheta^\top \Phi(x,\hat y)$ to decouple the soft-max from the regularization. 
$$
\min_{\scriptsize 
\begin{array}{c}\bftheta, \mu(x,\hat y) \\ \mu(x,\hat y) = \bftheta^\top \Phi(x,\hat y)\end{array}
} \left\{  \sum_{(x,y) \in \S} \epsilon \ln \sum_{\hat y} \exp \frac{e_y(\hat y) + \mu(x,\hat y)}{\epsilon}  - \bfd^\top \bftheta + \frac{C}{p} \|\bftheta\|_p^p \right\},
$$
To maintain consistency, we add the constraints $\mu(x,\hat y) = \bftheta^\top \Phi(x,\hat y)$, for every $(x,y) \in \S$ and every $\hat y \in \Y$. We compute the Lagrangian by adding the Lagrange multipliers $p_{x,y}(\hat y)$
$$L() = \sum_{(x,y) \in \S} \epsilon \ln \sum_{\hat y \in \Y} \exp \frac{e_y(\hat y) + \mu(x,\hat y)}{\epsilon} - \bfd^\top \bftheta + \frac{C}{p} \|\bftheta\|_p^p - \sum_{(x,y) \in \S, \hat y \in \Y} p_{x,y}(\hat y) \left( \mu(x,\hat y) - \bftheta^\top \Phi(x,\hat y) \right).$$
The dual function is a function of the Lagrange multipliers, and it is computed by minimizing the Lagrangian, namely $q(\bfp_{x,y}) = \min_{\bfmu,\bftheta} L(\bfmu,\bftheta,\bfp_{x,y})$. In particular the dual function can be written as 
$$\small \sum_{(x,y)} \min_{\mu(x,\hat y)} \left\{\epsilon \ln \sum_{\hat y} \exp \frac{e_y(\hat y) + \mu(x,\hat y)}{\epsilon} - \sum_{\hat y} \mu(x,\hat y) p_{x,y}(\hat y) \right\} + \min_{\bftheta} \left\{\frac{C}{p}\|\bftheta\|_p^p - \bftheta^\top (\hspace{-0.1cm} \sum_{(x,y), \hat y}  \hspace{-0.2cm} p_{x,y}(\hat y) \Phi(x,\hat y) - \bfd ) \right\}$$
and composed from the conjugate dual of the soft-max and the conjugate dual of the $\ell_p$ norm. Recall that the conjugate dual for the soft-max is the entropy barrier $\epsilon H(p_{x,y})$ over the set of probability distributions $\Delta_{\Y}$ (cf. \cite{Wainwright08} Theorem 8.1), and that the linear shift of the soft-max argument by $e_y(\hat y)$ result in the linear shift of the conjugate dual, thus we get the first part of the dual function $\sum_{(x,y)} (\epsilon H(p_{x,y}) + e_y^\top p_{x,y})$. Similarly, the conjugate dual of $\frac{1}{p}\|\bftheta\|_p^p$ is $\frac{1}{q}\|\bfz\|_q^q$ for the dual norm $1/p + 1/q = 1$ (cf. \cite{Rockafellar70}), where in our case $\bfz =  \sum_{(x,y), \hat y} p_{x,y}(\hat y) \Phi(x,\hat y) - \bfd$. \eop \\

\begin{theorem}
\label{theorem:approx}
The approximation of the structured prediction program in (3) takes the form
\begin{eqnarray*}
\small 
\min_{\lambda_{x,y,v \rightarrow \alpha}, \bftheta} && \sum_{(x,y) \in \S,v} \epsilon c_v \ln \sum_{\hat y_v} \exp \left( \frac{ e_y(\hat y_v) + \sum_{r: v \in V_{r,x}} \theta_r \phi_{r,v}(x,\hat y_v) -  \sum_{\alpha \in N(v)} \lambda_{x,y,v \rightarrow \alpha}(\hat y_v)}{\epsilon c_v} \right) \\ 
&& \hspace{-0.5cm} + \hspace{-0.3cm}  \sum_{(x,y) \in \S ,\alpha} \hspace{-0.3cm}  \epsilon c_\alpha \ln \sum_{\hat y_\alpha} \exp \left( \frac{\sum_{r: \alpha \in E_r} \theta_r \phi_{r,\alpha}(x,\hat y_\alpha)  + \sum_{v \in N(\alpha)} \lambda_{x,y,v \rightarrow \alpha}(\hat y_v)}{\epsilon c_\alpha} \right) - \bfd^\top \bftheta - \frac{C}{p} \left\| \bftheta \right\|_p^p  
\end{eqnarray*}
\end{theorem}
{\bf Proof:} We add auxiliary variables $\bfz$ and constrain them such that $$z_r = \sum_{(x,y) \in \S, v \in V_{r,x}, \hat y_v} b_{x,y,v}(\hat y_v) \phi_{r,v}(x,\hat y_v) + \sum_{(x,y) \in \S, \alpha \in E_{r,x}, \hat y_\alpha} b_{x,y,\alpha}(\hat y_\alpha) \phi_{r,\alpha}(x,\hat y_\alpha).$$ We derive the Lagrangian by introducing the Lagrange multipliers $\lambda_{x,y,v \rightarrow \alpha}(\hat y_v)$ for every marginalization constraint $\sum_{\hat y_\alpha \setminus \hat y_v} b_{x,y,\alpha}(\hat y_\alpha) = b_{x,y,v}(\hat y_v)$, and Lagrange multipliers $\theta_r$ for every equality constraint involving $z_r$. In particular, the Lagrangian has the form: 
\bea
L() &=&  \sum_{(x,y) \in \S} \left(\sum_{\alpha \in E} \epsilon  c_\alpha H(b_{x,y,\alpha}) + \sum_{v \in V} \epsilon  c_v H(b_{x,y,v})  + \sum_{v \in V, \hat y_v} b_{x,y,v}(\hat y_v) e_{y,v}(\hat y_v) \right) - \frac{C^{1-q}}{q}  \| \bfz - \bfd \|_q^q \\
&& + \sum_r \theta_r \left( \sum_{(x,y) \in \S ,v \in V_r, \hat y_v} b_{x,y,v}(\hat y_v) \phi_{r,v}(x,\hat y_v) + \sum_{(x,y) \in \S ,\alpha \in E_r, \hat y_\alpha} b_{x,y,\alpha}(\hat y_\alpha) \phi_{r,\alpha}(x,\hat y_\alpha)  - z_r \right) \\
&& + \sum_{v,\alpha \in N(v),\hat y_v} \lambda_{x,y,v\rightarrow \alpha}(\hat y_v) \left( \sum_{\hat y_\alpha \setminus \hat y_v} b_{x,y,\alpha}(\hat y_\alpha) - b_{x,y,v}(\hat y_v) \right) \\
\eea
We obtain the dual function by minimizing the beliefs over their compact domain, i.e. $$q(\bflambda_{x,y,v \rightarrow \alpha}, \bftheta) = \max_{b_{x,y,v}(\hat y_v) \in \Delta_{\Y_v}, \;\;\;  b_{x,y,\alpha}(\hat y_\alpha) \in \Delta_{\Y_\alpha}} L(\bfb_{x,y,v},\bfb_{x,y,\alpha},\bflambda_{x,y,v \rightarrow \alpha}, \bftheta),$$
Deriving the dual by minimizing over the compact set of beliefs enables us to obtain an {\em unconstrained} dual, which corresponds to the approximated structured prediction program. The dual function is described by the conjugate dual function:
\bea
&& \hspace{-0.7cm} \sum_{(x,y) \in \S,v} \max_{b_{x,y,v} \in \Delta_{\Y_v}} \left\{ \epsilon c_v H(b_{x,y,v}) + \sum_{\hat y_v} b_{x,y,v}(\hat y_v) \left( e_y(\hat y_v) + \sum_{r: v \in V_r} \theta_r \phi_{r,v}(x,\hat y_v) -\sum_{\alpha \in N(v)} \lambda_{x,y,v \rightarrow \alpha} (\hat y_v)  \right) \right\} \\
&& + \sum_{(x,y) \in \S,\alpha} \max_{b_{x,y,\alpha} \in \Delta_{\Y_\alpha}} \left\{ \epsilon c_\alpha H(b_{x,y,\alpha}) + \sum_{\hat y_\alpha} b_{x,y,\alpha}(\hat y_\alpha) \left(\sum_{r: \alpha \in E_r} \theta_r \phi_{r,\alpha}(x,\hat y_\alpha) + \sum_{v \in N(\alpha)} \lambda_{x,y,v \rightarrow \alpha} (\hat y_v) \right) \right\} \\
&&+ \max_{\bfz} \left\{ - \frac{C^{1-q}}{q}  \| \bfz - \bfd \|_q^q - \bfz^\top \bftheta \right\}
\eea
Its final form is derived similarly to Claim \ref{claim:dual}, where we show that the conjugate dual of the entropy barrier is the soft-max function and the conjugate dual of the $\ell_q^q$ is the $\ell_p^p$. \eop

\begin{lemma}
\label{lemma:lambda}
Given a vertex $v$ in the graphical model, the optimal $\lambda_{x,y,v \rightarrow \alpha}(\hat y_v)$ for every $\alpha \in N(v), \hat y_v \in \Y_v, (x,y) \in \S$ in the approximated program of Theorem \ref{theorem:approx} satisfies
\begin{eqnarray*}
 \mu_{x,y,\alpha\rightarrow v}(\hat y_v) &=& \epsilon c_\alpha \ln \left(\sum_{\hat y_\alpha \setminus \hat y_v} \exp \left( \frac{\sum_{r: \alpha \in E_{r,x}} \theta_r \phi_{r,\alpha}(x,\hat y_\alpha) + \sum_{u \in N(\alpha) \setminus v} \lambda_{x,y,u \rightarrow \alpha}(\hat y_u)}{\epsilon c_\alpha} \right) \right)  \\
 \lambda_{x,y,v\rightarrow\alpha}(\hat y_v) &=&  \frac{c_\alpha}{\hat c_v} \left(e_{y,v}(\hat y_v) + \sum_{r: v \in V_{r,x}} \theta_r \phi_{r,v}(x,\hat y_r) + \sum_{\beta\in N(v)} \mu_{x,y,\beta\rightarrow v}(\hat y_v)\right) - \mu_{x,y,\alpha\rightarrow v}(\hat y_v) + c_{x,y,v \rightarrow \alpha}
\end{eqnarray*}
for every constant $c_{x,y,v \rightarrow \alpha}$\footnote{For numerical stability in our algorithm we set $c_{x,y,v \rightarrow \alpha}$ such that $\sum_{\hat y_v} \lambda_{x,y,v \rightarrow \alpha}(\hat y_v)=0$}, where $\hat c_v = c_v + \sum_{\alpha \in N(v)} c_\alpha$. In particular, if either $\epsilon$ and/or $c_\alpha$ are zero then $\mu_{x,y,\alpha \rightarrow v}$ corresponds to the $\ell_\infty$ norm and can be computed by the max-function. 
Moreover, if either $\epsilon$ and/or $c_\alpha$ are zero in the objective, then the optimal $ \lambda_{x,y,v\rightarrow\alpha}$ can be computed for any arbitrary $c_\alpha >0 $, similarly for $c_v>0$.
\end{lemma}
{\bf Proof:} For a given $x,y$ and $v$, optimizing $\lambda_{x,y,v \rightarrow \alpha}(\hat y_v)$ for every $\alpha \in N(v)$ and $\hat y_v \in \Y_v$ while holding the rest of the variables fixed, reduces the problem to
\begin{eqnarray*}
\small 
\min_{\lambda_{x,y,v \rightarrow \alpha}(\hat y_v)} && \epsilon c_v \ln \sum_{\hat y_v} \exp \left( \frac{ e_y(\hat y_v) + \sum_{r: v \in V_{r,x}} \theta_r \phi_{r,v}(x,\hat y_v) -  \sum_{\alpha \in N(v)} \lambda_{x,y,v \rightarrow \alpha}(\hat y_v)}{\epsilon c_v} \right) \\ 
&& \hspace{-0.5cm} + \hspace{-0.3cm}  \sum_{\alpha \in N(v)} \hspace{-0.3cm}  \epsilon c_\alpha \ln \sum_{\hat y_\alpha} \exp \left( \frac{\sum_{r: \alpha \in E_r} \theta_r \phi_{r,\alpha}(x,\hat y_\alpha)  + \sum_{v \in N(\alpha)} \lambda_{x,y,v \rightarrow \alpha}(\hat y_v)}{\epsilon c_\alpha} \right)
\end{eqnarray*}
Let $$\mu_{x,y,\alpha \rightarrow v}(\hat y_v) = c_\alpha \ln \sum_{\hat y_\alpha \setminus \hat y_v} \exp \left( \frac{\sum_{r: \alpha \in E_r} \theta_r \phi_{r,\alpha}(x,\hat y_\alpha)  + \sum_{u \in N(\alpha) \setminus v} \lambda_{x,y,u \rightarrow \alpha}(\hat y_u)}{\epsilon c_\alpha} \right),$$ and also $\phi_{x,y,v}(\hat y_v) = e_y(\hat y_v) + \sum_{r: v \in V_{r,x}} \theta_r \phi_{r,v}(x,\hat y_v)$. We find the optimal $\lambda_{x,y,v \rightarrow \alpha}(\hat y_v)$ whenever the gradient vanishes, i.e. 
$$ 
0 = \nabla \left\{\epsilon c_\alpha \ln \sum_{\hat y_v} \exp \left(\frac{\mu_{x,y,\alpha \rightarrow v}(\hat y_v) + \lambda_{x,y,v \rightarrow \alpha}(\hat y_v)}{\epsilon c_\alpha}\right) + \epsilon c_v \ln \sum_{\hat y_v} \exp \left( \frac{ \phi_{x,y,v}(\hat y_v) -  \sum_{\alpha \in N(v)} \lambda_{x,y,v \rightarrow \alpha}(\hat y_v)}{\epsilon c_v} \right) \right\} 
$$
Taking the vanishing point of the gradient we derive two probabilities over $\hat y_v$ that need to be the same, namely $$\frac{\exp \left(\frac{\mu_{x,y,\alpha \rightarrow v}(\hat y_v)  + \lambda_{x,y,v \rightarrow \alpha}(\hat y_v) }{\epsilon c_\alpha} \right)}{\sum_{\tilde y_v} \exp \left(\frac{\mu_{x,y,\alpha \rightarrow v}(\tilde y_v) + \lambda_{x,y,v \rightarrow \alpha}(\tilde y_v) }{\epsilon c_\alpha} \right)} = \frac{\exp \left( \frac{ \phi_{x,y,v}(\hat y_v) -  \sum_{\beta \in N(v)} \lambda_{x,y,v \rightarrow \beta}(\hat y_v)}{\epsilon c_v} \right)}{\sum_{\tilde y_v} \exp \left( \frac{ \phi_{x,y,v}(\tilde y_v) -  \sum_{\beta \in N(v)} \lambda_{x,y,v \rightarrow \beta}(\tilde y_v)}{\epsilon c_v} \right)}. 
$$
For simplicity we need to consider only the numerator, while taking one degree of freedom in the normalization. Taking log of the numerator we get that the gradient vanishes if the following holds 
\be
\label{eq:opt-lambda}
\hat c_{x,y,v\rightarrow \alpha} + \frac{\mu_{x,y,\alpha \rightarrow v}(\hat y_v) + \lambda_{x,y,v \rightarrow \alpha}(\hat y_v) }{c_\alpha}  = \frac{ \phi_{x,y,v}(\hat y_v) -  \sum_{\beta \in N(v)} \lambda_{x,y,v \rightarrow \beta}(\hat y_v)}{c_v}.
\ee
Multiplying both sides of the equation by $c_v c_\alpha$, and summing both sides with respect to $\beta \in N(v)$ gives
\be
\label{eq:opt-sum}
\tilde c_{x,y,v\rightarrow \alpha} + c_v \sum_{\beta \in N(v)} \left(\mu_{x,y,\beta \rightarrow v}(\hat y_v) + \lambda_{x,y,v \rightarrow \beta}(\hat y_v) \right) = \left(\sum_{\beta \in N(v)} c_\beta \right) \left(\phi_{x,y,v}(\hat y_v) -  \sum_{\beta \in N(v)} \lambda_{x,y,v \rightarrow \beta}(\hat y_v) \right).
\ee 
We wish to find the optimal value of $\lambda_{x,y,v \rightarrow \alpha}(\hat y_v)$, namely the value that satisfies Eq. (\ref{eq:opt-lambda}). For that purpose we recover the value of $\sum_{b \in N(v)} \lambda_{x,y,v \rightarrow \beta}(\hat y_v)$ from (\ref{eq:opt-sum}): 
$$
\tilde c_{x,y,v\rightarrow \alpha} + \left(c_v + \sum_{\beta \in N(v)} c_\beta \right) \left(\sum_{\beta \in N(v)} \lambda_{x,y,v \rightarrow \beta}(\hat y_v) \right) = \left(\sum_{\beta \in N(v)} c_\beta \right) \phi_{x,y,v}(\hat y_v) - c_v \sum_{\beta \in N(v)} \mu_{x,y,\beta \rightarrow v}(\hat y_v).
$$
Plugging this into   \ref{eq:opt-lambda} gives
$$
\mu_{x,y,\alpha \rightarrow v}(\hat y_v) + \lambda_{x,y,v \rightarrow \alpha}(\hat y_v) = \frac{c_\alpha}{c_v + \sum_{\beta \in N(v)} c_\beta} \left( \phi_{x,y,v}(\hat y_v) + \sum_{\beta \in N(v)} \mu_{x,y,\beta \rightarrow v}(\hat y_v)\right) + c_{x,y,v \rightarrow \alpha}
$$ which concludes the proof for $\epsilon,c_\alpha,c_v > 0$. Whenever any of these quantitates is zero, Danskin's theorem (cf. \cite{Bertesekas03}, Theorem 4.5.1) states that its corresponding subgradient is described by a probability distribution over its maximal assignments. Therefore if $c_\alpha=0$ in the objective function, then equality (\ref{eq:opt-lambda}) holds for every $c_\alpha$, and similarly whenever $c_v=0$ in the objective, equality holds for every $c_v$.    
\eop

\begin{lemma}
\label{lemma:theta}
The gradient of the approximated structured prediction program in Theorem \ref{theorem:approx} with respect to $\theta_r$ equals to 
$$
 \sum_{(x,y) \in \S,v \in V_{r,x},\hat y_v}  \hspace{-0.5cm}  b_{x,y,v}(\hat y_v) \phi_{r,v}(x,\hat y_v) +   \hspace{-0.5cm}  \sum_{(x,y) \in \S ,\alpha \in E_{r,x},\hat y_\alpha} \hspace{-0.5cm} b_{x,y,\alpha}(\hat y_\alpha) \phi_{r,\alpha}(x,\hat y_\alpha) - d_r + C \cdot |\theta_r|^{p-1} \cdot \mbox{sign}(\theta_r), 
$$
where 
\begin{eqnarray*}
b_{x,y,v}(\hat y_v) &\propto& \exp\left(\frac{e_y(\hat y_v) + \sum_{r: v \in V_{r,x}} \theta_r \phi_{r,v}(x,\hat y_v) - \sum_{\alpha \in N(v)} \lambda_{x,y,v \rightarrow \alpha}(\hat y_v)}{\epsilon c_v} \right) \\
b_{x,y,\alpha}(\hat y_\alpha) &\propto& \exp\left(\frac{\sum_{r: \alpha \in E_{r,x}} \theta_r \phi_{r,\alpha}(x,\hat y_\alpha) + \sum_{v \in N(\alpha)} \lambda_{x,y,v \rightarrow \alpha}(\hat y_\alpha)}{\epsilon c_\alpha} \right) \\
\end{eqnarray*}
However, if either $\epsilon$ and/or $c_\alpha$ equal zero, then the beliefs $ b_{x,y,\alpha}(\hat y_\alpha)$ can be taken from the set of probability distributions over support of the max-beliefs, namely $b_{x,y,\alpha}(\hat y^*_\alpha)>0$ only if $\hat y^*_\alpha \in \mbox{argmax}_{\hat y_\alpha} \left\{ \sum_{r: \alpha \in E_{r,x}} \theta_r \phi_{r,\alpha}(x,\hat y_\alpha) + \sum_{v \in N(\alpha)} \lambda_{x,y,v \rightarrow \alpha}(\hat y_\alpha) \right\}$. Similarly for $b_{x,y,v}(\hat y^*_v)$ whenever $\epsilon$ and/or $c_v$ equal zero. 
\end{lemma} 
 {\bf Proof:} This is a direct computation of the gradient. In the special case of $\epsilon,c_\alpha = 0$ then $b_{x,y,\alpha}(\hat y_\alpha)$ corresponds to the subgradient and similarly when $\epsilon,c_v=0$, cf. Danskin's theorem (cf. \cite{Bertesekas03}, Theorem 4.5.1). \eop
 
\begin{claim}
\label{claim:conv}
The block coordinate descent algorithm in lemmas \ref{lemma:lambda} and \ref{lemma:theta} monotonically reduces the approximated structured prediction objective in Theorem \ref{theorem:approx}, therefore the value of its objective is guaranteed to converge. Moreover, if $\epsilon,c_\alpha,c_v > 0$, the objective is guaranteed to converge to the global minimum, and its sequence of beliefs are guaranteed to converge to the unique solution of the approximated structured prediction dual.  
\end{claim}
{\bf Proof:} The approximated structured prediction dual  is strictly concave in the dual variables $b_{x,y,v}(\hat y_v), b_{x,y,\alpha}(\hat y_\alpha),\bfz$ subject to linear constraints. The claim properties are a direct consequence of \cite{Tseng87} for this type of programs. \eop

\begin{claim}
\label{claim:non-conv}
Whenever the approximated structured prediction is non convex, i.e., $\epsilon, c_\alpha > 0$ and $c_v < 0$, the algorithm in lemmas \ref{lemma:lambda} and \ref{lemma:theta} is not guaranteed to converge, but whenever it converges it reaches a stationary point of the primal and dual approximated structured prediction programs. 
\end{claim}
{\bf Proof:} 
The approximated structured prediction in Theorem \ref{theorem:approx} is unconstrained. The update rules defined in Lemmas \ref{lemma:lambda} and \ref{lemma:theta} are directly related to vanishing points of the gradient of this function, even when it is non-convex. Therefore a stationary point of the algorithm corresponds to an assignment $\lambda_{x,y,v \rightarrow \alpha}(\hat y_v), \theta_r$ for which the gradient equals zero, or equivalently a stationary point of the approximated structured prediction.

The dual approximated structured prediction in (\ref{eq:dual-approx}) is a constrained optimization and its stationary points are saddle points of the Lagrangian, defined in Theorem \ref{theorem:approx}, with respect to the probability simplex $b_{x,y,v}(\hat y_v) \in \Delta_{\Y_v}$ and $b_{x,y,\alpha}(\hat y_\alpha)  \in \Delta_{\Y_\alpha}$. Note that since $\epsilon,c_\alpha,c_v \ne 0$ the entropy functions act as barrier functions on the nonnegative cone, therefore we need not consider the nonnegative constraints over the beliefs. In the following we show that at stationary points the inferred beliefs of the Lagrangian satisfy the marginalization constraints, therefore are saddle points of the Lagrangian.  

When $\epsilon,c_\alpha > 0$ the stationary beliefs $b_{x,y,\alpha}(\hat y_\alpha)$ are achieved by maximizing over $\Delta_{\Y_\alpha}$, resulting in 
$$b_{x,y,\alpha}(\hat y_\alpha) \propto \exp\left(\frac{\sum_{r: \alpha \in E_{r,x}} \theta_r \phi_{r,\alpha}(x,\hat y_\alpha) + \sum_{v \in N(\alpha)} \lambda_{x,y,v \rightarrow \alpha}(\hat y_\alpha)}{\epsilon c_\alpha} \right).$$ However, since $c_v < 0$ the stationary beliefs $b_{x,y,v}(\hat y_v)$ are achieved by {\em minimizing} over $\Delta_{\Y_v}$ resulting in $$b_{x,y,v}(\hat y_v) \propto \exp\left(\frac{e_y(\hat y_v) + \sum_{r: v \in V_{r,x}} \theta_r \phi_{r,v}(x,\hat y_v) - \sum_{\alpha \in N(v)} \lambda_{x,y,v \rightarrow \alpha}(\hat y_v)}{\epsilon c_v} \right).$$ To prove these beliefs correspond to a stationary point we show that they satisfy the marginalization constraints. This fact is a direct consequence of the update rule in Lemma \ref{lemma:lambda}, where by direct computation one can verify that $$\sum_{\hat y_\alpha \setminus \hat y_v} b_{x,y,\alpha}(\hat y_\alpha) \propto \exp\left( \frac{\mu_{x,y,\alpha \rightarrow v}(\hat y_v) + \lambda_{x,y,v \rightarrow \alpha}(\hat y_v)}{\epsilon c_\alpha} \right).$$ Following the definition of $b_{x,y,v}(\hat y_v)$ one can see that the update rule in Lemma \ref{lemma:lambda} enforces the marginalization constraints. This implies that the gradient of the approximated structured prediction program measures the disagreements between $\sum_{\hat y_\alpha \setminus \hat y_v} b_{x,y,\alpha}(\hat y_\alpha)$ and $b_{x,y,v}(\hat y_v)$, and the gradient vanishes only when they agree. Therefore these beliefs correspond to a saddle point of the Lagrangian.
\eop \\

\vspace{-0.2cm}
\small{
\bibliographystyle{plain}
\bibliography{approx-sup.bib}

\begin{thebibliography}{1}

\bibitem{Bertesekas03}
D.~P. Bertsekas, A.~Nedi\'{c}, and A.~E. Ozdaglar.
\newblock {\em Convex Analysis and Optimization}.
\newblock Athena Scientific, 2003.

\bibitem{Rockafellar70}
R.T. Rockafellar.
\newblock {\em {Convex analysis}}.
\newblock Princeton university press, 1970.

\bibitem{Tseng87}
P.~Tseng and D.P. Bertsekas.
\newblock {Relaxation methods for problems with strictly convex separable costs
  and linear constraints}.
\newblock {\em Mathematical Programming}, 38(3):303--321, 1987.

\bibitem{Wainwright08}
M.J. Wainwright and M.I. Jordan.
\newblock {Graphical models, exponential families, and variational inference}.
\newblock {\em Foundations and Trends{\textregistered} in Machine Learning},
  1(1-2):1--305, 2008.

\end{thebibliography}
}

\end{document}